\RequirePackage{amsmath}
\documentclass[a4paper]{llncs}

\usepackage{subfigure}
\usepackage{amssymb}
\usepackage{graphicx}
\usepackage{xcolor}
\usepackage[normalem]{ulem}
\usepackage[linesnumbered,ruled]{algorithm2e} %Pseudocode
\usepackage{floatrow}
\usepackage[
  separate-uncertainty = true,
  multi-part-units = repeat
]{siunitx}

\newcommand{\todo}[1]{\textcolor{red}{{#1}}}  % todo 

\begin{document}

\title{Iterative Interaction Training for Segmentation Editing Networks}
%\title{Efficient editing of automatic segmentations by iterative training}
%
% hybrid approach which combines automatic and interactive CNNs for multi-class segmentation
% AuxiSegNet - Auxillary Segmentation Network
% Combining Automatic and Interactive Neural Networks for Multi-Class Segmentation
% Robust editing 
% Efficient editing of automatic segmentations by iterative training
%
\titlerunning{Iterative Interaction Training for Segmentation Editing Networks}

\author{Gustav Bredell, Christine Tanner, Ender Konukoglu}
\authorrunning{}
\institute{Computer Vision Laboratory, ETH Zurich, Zurich, Switzerland
\textbf{gbredell@student.ethz.ch}}

\maketitle              % typeset the title of the contribution

\begin{abstract}
Automatic segmentation has great potential to facilitate morphological measurements while simultaneously increasing efficiency. Nevertheless often users want to edit the segmentation to their own needs and will need different tools for this. There has been methods developed to edit segmentations of automatic methods based on the user input, primarily for binary segmentations. Here however, we present an unique training strategy for convolutional neural networks (CNNs) trained on top of an automatic method to enable interactive segmentation editing that is not limited to binary segmentation. By utilizing a robot-user during training, we closely mimic realistic use cases to achieve optimal editing performance. In addition, we show that an increase of the iterative interactions during the training process up to ten improves the segmentation editing performance substantially. Furthermore, we compare our segmentation editing CNN (interCNN) to state-of-the-art interactive segmentation algorithms and show a superior or on par performance.
\end{abstract}

\section{Introduction}

Segmentation is one of the main medical image analysis tasks that when automated substantially facilitates morphological measurements and increase efficiency in treatment planning~\cite{pasquier2007automatic,toth2011accurate,vos2012automatic}. With the introduction of machine learning and especially convolutional neural networks (CNNs) the performance of automatic segmentation approaches improved greatly~\cite{litjens2017survey}. Recent studies showed that CNN-based approaches were able to achieve inter- and intra-expert performance in certain segmentation tasks, for example prostate segmentation in Magnetic Resonance Images (MRIs) as shown in~\cite{litjens2014evaluation,vanGinneken2018PROMISE12}. 
% https://promise12.grand-challenge.org/evaluation/results
% results over 85 are better than the second observer
% tian2018psnet  winner ISBI13 challenge, but no inter-observer variability evaluated
Although these approaches achieve impressive performance on average, when considering an individual image, there are often parts of the segmentation users would like to change and improve to fit their needs. The need for edits and improvement is even larger when the test image differs slightly from the training dataset, for example due to scanner differences, and more errors are expected.

To address the need for editing, interactive segmentation algorithms have been proposed such as GrabCut, GeoS or Random Walker~\cite{rother2004grabcut,criminisi2008geos,grady2005random} that allow operators to modify segmentations. Even though accurate results have been shown with these methods, the interaction can be time consuming as large number of interactions might be necessary. In particular, updates aiming to correct segmentation in one region can lead to inaccuracies in another region, consequently requiring further interactions.

In recent years, studies such as~\cite{amrehn2017ui,mahadevan2018iteratively,wang2018interactive} proposed CNNs for interactive segmentations and showed better results compared to traditional methods. These initial works focused on segmenting objects in medical images from scratch using simple user interactions, and mostly in the form of binary segmentations. 
More recently, authors in~\cite{wang2018deepigeos} proposed a CNN-based method for editing segmentations predicted by an automatic algorithm, one of the most important steps in translating automatic segmentations in practice, and showed the benefits for binary segmentations. 
In the same work, authors assumed multiple scribbles to be made at a single time and the editing network was trained to take into account all the edits, initial prediction and the image to generate an updated segmentation. 
This training strategy may not be ideal since it does not take into account the fact that a user may be interacting with the tool over several iterations, each time providing scribbles based on the result of the last update. 

In this work, we present a different strategy for training a CNN that interactively edits segmentations. As in \cite{wang2018deepigeos}, we assume the editing CNN is an auxiliary tool that supports a base segmentation algorithm and is optimized to take into account user edits and improve segmentation accuracy. Different than \cite{wang2018deepigeos}, we investigate training in an iterative interaction fashion on simulated user inputs and we also focus on multi-label segmentation problems as well as binary ones. 
We assess the potential of the proposed training strategy on the prostate data of the NCI-ISBI 2013 challenge and show the value of iterative interaction training. 
Moreover, we empirically compare networks for editing segmentations with a state-of-the-art fully interactive segmentation algorithm that segments the image from scratch using user-made scribbles. 

\section{Methods}
\begin{figure}[tbh]
\centering
\includegraphics[width=0.85\textwidth]{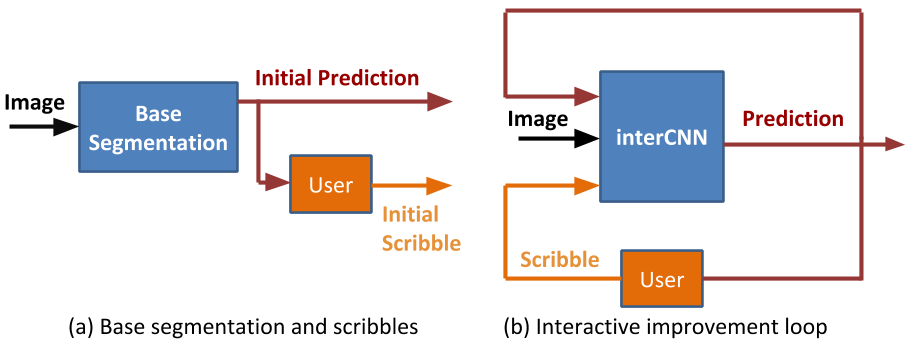}
\caption{Illustration of interactive segmentation editing networks. (a) generation of initial prediction with base segmentation and first user input, (b) interactive improvement loop with proposed interCNN. Here, we use a CNN for the base segmentation algorithm for demonstration but other methods can be used. interCNN can be applied iteratively until the segmentation is satisfactory. During training, to make it feasible, the user is replaced by a robot user that places scribbles based on the discrepancy between ground truth and predicted segmentations for the training images.}
\label{fig:framework}
\end{figure}

Interactive segmentation editing networks, which we refer to as \emph{interCNN}, are trained on top of a base segmentation algorithm, specifically to interpret user inputs and make appropriate adjustments to the predictions of the base algorithm. 
During test time, an interCNN sees the image, initial predictions of the base algorithm and user edits in the form of scribbles, and combines all to create a new segmentation, see Figure~\ref{fig:framework}. 
In case the new segmentation needs more edits, an interCNN can be applied in an iterative fashion until the segmentation is satisfactory by accepting additional scribbles and taking the image and its own predictions as inputs. 
\begin{algorithm}[tbh]
\DontPrintSemicolon
\caption{training interCNN for $B$ batch and $K$ interaction iterations\label{alg:interCNNtraining}\;}
%\caption{training of interCNN for the total number of $B$ batch iterations and $K$ interaction iterations\label{alg:interCNNtraining}\;}
\SetKwInOut{Input}{Input}
\SetKwInOut{Output}{Output}
\Input{images $\mathbf{I}^b$, ground-truth labels $\mathbf{L}^b$}
%, predictions $\mathbf{P}^b$, scribble masks $\mathbf{S}^b$}
\Output{interCNN weights $\mathbf{W}_K$, predictions $\mathbf{P}^b_K$}
\For{$b \in \{1,2,...,B\}$}
{
$\mathbf{P}^b_0 \leftarrow $ autoCNN($\mathbf{I}^b$)\;
$\mathbf{S}^b_0 \leftarrow $ random-user($\mathbf{P}^b_0,\mathbf{L}^b$)\;
\For{$k \in \{1,...,K\}$}
{
$\mathbf{P}^b_{k} \leftarrow $ interCNN($\mathbf{I}^b,\mathbf{P}^b_{k-1},\mathbf{S}^b_{k-1}$)\; 
$\mathbf{S}^b_{k}  \leftarrow$ random-user($\mathbf{P}^b_{k},\mathbf{L}^b$)\;
backpropagate cross-entropy($\mathbf{P}^b_{k}$,$\mathbf{L}^b$) loss to update $\mathbf{W}_{k}$\;
}
}
\end{algorithm}
Training of an interCNN can be done in two ways. First, as done in~\cite{wang2018deepigeos}, given the segmentation of the base network, a set of scribbles are provided and the interCNN is trained to update the segmentation the best way possible by using all the scribbles, image and the base network's segmentation. 
Ideally, human users should provide the scribbles during the training, however, this is clearly infeasible and a robot user is often utilized to provide the scribbles and has been shown to perform well. 
\paragraph{{\bf Iterative interaction training:}} The alternative training strategy, which we propose here, is to replicate the testing procedure and integrate iterative interactions to optimize the network. An overview of this strategy is presented as a pseudocode in Algorithm~\ref{alg:interCNNtraining}.
Images in the training set ($\mathbf{I}^b$) are fed batch-wise into the base algorithm to create initial predictions ($\mathbf{P}^b_0$). Scribbles ($\mathbf{S}^b_k$) are produced by a robot user based on the discrepancy between $\mathbf{P}^b_k$ and the ground truth segmentations ($\mathbf{L}^b$). $\mathbf{S}^b_k$ has an image format in which the user-selected wrongly classified pixels are marked according to their correct class and all other pixels are set to 
$\max({\mathbf{L}^b}$$)+$$1$. The initial scribbles $\mathbf{S}^b_0$, along with $\mathbf{P}^b_0$ and $\mathbf{I}^b$ are subsequently fed into interCNN to get an updated prediction ($\mathbf{P}^b_{1}$). Based on $\mathbf{P}^b_{1}$ new scribbles are produced by the robot user ($\mathbf{S}^b_{1}$) and are fed into the interCNN in the next iteration ($k$+1) together with $\mathbf{I}^b$ and $\mathbf{P}^b_{1}$. During interaction iteration $k$ the weights of interCNN ($\mathbf{W}_k$) are updated with backpropagation based on the cross-entropy loss between 
$\mathbf{P}^b_{k}$ and $\mathbf{L}^b$. This is repeated for a fixed number of $K$ interaction iterations before moving on to the next batch of images $\mathbf{I}^{b+1}$. 
\paragraph{{\bf Base segmentation method:}} Ideally, the base segmentation algorithm is arbitrary. An interCNN can be used with any algorithm. In this work, we used a segmentation CNN as the base algorithm due to their superior performance, similar to \cite{wang2018deepigeos}, and refer to it as autoCNN. 
%For both the autoCNN and the interCNN a U-Net architecture was used~\cite{ronneberger2015u}. 
\paragraph{{\bf Network architecture:}}
We used a U-Net architecture~\cite{ronneberger2015u} for both the autoCNN and interCNN. 
It has been shown that this architecture produces automatic segmentation results on medical images that is comparable to more complex architectures~\cite{tian2018psnet,zhu2017deeply}. 
Our implementation consists of 4 down- and 4~up-convolutional layers.
Each down-convolutional layer is also connected to its respective up-convolutional layer through skip-connections. 
The final prediction of the U-Net is obtained by a softmax layer. 
The input consisted of 320$\times$320 pixel patches. 
Most U-Net networks take the image as the only input. 
interCNN, however, takes three inputs: image, prediction and scribble mask.
%, which we will explain next. 

For the base segmentation model autoCNN, we also used the same U-Net architecture but with only the image as the input. For both interCNN and autoCNN, we used drop-out and batch normalization during training~\cite{srivastava2014dropout,ioffe2015batch}.

We note that more complex networks can also be used both for autoCNN and interCNN. Here, we use a relatively simple architecture since our focus is on the training strategy rather than the architecture. 

\paragraph{{\bf Robot user:}} The robot user we utilized for training the network is based on the model introduced by Nickisch et al.~\cite{nickisch2010learning}. Here a random-user model is used. 
%In each iteration a scribble is produced for every class in the image by comparing the prediction to the ground truth and all the incorrectly classified pixels are identified. Subsequently, a pixel from the incorrectly classified pixels is chosen randomly for each class respectively. 
At each iteration a scribble is produced for every class in the image by comparing the prediction to the ground truth. First, all incorrectly classified pixel are identified. Subsequently, a pixel from the incorrectly classified pixels is chosen randomly for each class separately. 
In a next step, a region of 9$\times$9 pixels is placed around each randomly chosen pixel and all the pixels in this region belonging to the class the scribble is currently made for are saved as the scribble for the respective class. 
%All the pixels in this region belonging to the \addct{same class as the center pixel}\chct{class the scribble is currently made for,} are saved as scribble for the respective class. 
This process is repeated for all classes in each iteration. The scribbles from all classes are then added together to obtain the final scribble mask for the respective iteration.
The randomness in choosing the scribbles prevents the interCNN from over-fitting to a specific strategy that the user may not reproduce during test time, for instance always choosing the center of gravity of the difference set. 
%there is no clear strategy behind selecting the scribble input and prevents the interCNN from overfitting to a specific strategy. 

%\subsection{Iterative Training}

%The training process of the framework is divided into two parts. First the autoCNN has to be trained on the training data. The goal of the first part is to obtain a CNN that is performing optimal with regard to automatic segmentation. After the autoCNN is trained, its parameters are frozen. In the second step of the training process interCNN is trained. For every batch, autoCNN produces an initial prediction, which is used to also produce initial scribbles. 
%The proposed method is trained given a base segmentation network. 
%The initial scribbles, initial prediction and image are then fed to the interCNN in a first iteration. The prediction interCNN produces with this input is then compared to the ground truth through the loss function and its parameters updated accordingly. Subsequently, the prediction produced by interCNN is used to make new scribbles. In a second iteration, these new scribbles, prediction from the first iteration and the original image is fed back to the interCNN and the network is again updated depending on the loss of its prediction. This process is shown in Fig.~\ref{fig:framework} and is continued for a fixed amount of iterations during training. \todo{[CT: very similar to text at beginning of this section. Can beginning be shortened?]}

%\subsection{Implementation Details}

\paragraph{{\bf Implementation details:}} We used PyTorch~\cite{paszke2017automatic} and Python to implement our U-net and robot user, respectively. The training took place on the in-house GPU cluster mainly consisting of GeForce GTX TITAN X with 12GB memory. The Adam optimization algorithm was used for training. The batch size was fixed to 4 images, learning rate to 0.0001 and the maximal number of iterations was 140'000. The images were normalized by taking the median of the training images and dividing all images by this value. To prevent over-fitting, data augmentation was used during training. 
For each batch, cropping, rotation or flipping was applied to all the images within the batch with a probability of 0.5. 
%Data augmentation was used to prevent over-fitting. 
%After each random batch sampling the following actions were applied to the batch images each with a probability of 0.5:  cropping, rotation and flipping. 

\section{Experiments and Results}
\label{sec:results}

%\subsection{Materials}
\paragraph{{\bf Materials:}}
%For our experiments w
We used the prostate dataset of the NCI-ISBI 2013 challenge~\cite{bloch2015nci}. The dataset consists of T2-weighted MRIs of the prostate acquired with a 3.0~T scanner. 
In total the dataset includes 60 patient volumes, each containing 15-20 slices. Of the 60 patients only 29 had multi-class ground truth segmentations, where the central gland and the peripheral zone were labeled. 
We focused our experiments on these 29 subjects to present results in multi-class segmentation. 
%As we expand the scope of interactive segmentation to multi-class in this work we only considered these 29 patients.

We randomly divided the patients into 4 groups G1-G4. 
G1 contained 15~patients and was used as training data for the base segmentation algorithm, autoCNN.
G2 consisted of 8 patients and was used as validation data for autoCNN. 
For training interCNN, both G1 and G2 
%(23 subjects) 
were used. 
Training interCNN with G2 is crucial, since often the base method already performs very well on its training data, so interCNN would not encounter large incorrect classifications if only trained with G1. 
%and at the same time served as additional training data for the interCNN. The training data for interCNN was thus \addct{G1 and G2}\chct{group one and group two} combined summing to 23 patients. The additional training data for interCNN is crucial, since often the autoCNN will already perform very well on its training data and will not allow interCNN to learn from large incorrect classifications. 
%In addition, o
One patient, G3, was used as validation data to select the best performing interCNN.
%Lastly, 
G4 constituted the test data and consisted of 5 patients.

For the benchmarking against other approaches, which were all focused on binary segmentation, we kept the same groups, but transformed the multi-class labels to binary by fusing the central gland and peripheral zone. 
\paragraph{{\bf Evaluation:}}
We employed the random robot-user for assessing test performance for the sake of efficiency. This neglects potential user errors but it does not simulate an ideal user nor favours a particular behaviour due to the randomness. 

The segmentation performance was quantified using the Dice score (DSC):
$DSC=\frac{2|S_g\cap{}S_p|}{|S_g|+|S_p|}$
where $S_g$ is the ground truth and $S_p$ is the predicted segmentation, and $|\cdot|$ denotes the number of pixels. We simulated that the user was interactively editing the proposed segmentations of each test image up to 20 times. We calculated the Dice score after every simulated
%random robot-user 
user interaction to see how the segmentation results are influenced by the number of user inputs.

\paragraph{{\bf Computation speed:}} The interCNN produced an updated prediction per interactive iteration with a mean time of $\SI{3.9 \pm 0.2}{\ms}$, thus enabling real-time use. GrabCut needs 1.2s per update (openCV implementation). Hence a substantial increase in update speed is obtained with interCNN over GrabCut.
%traditional methods.}

%\subsection{Iteration Training Parameter}
\paragraph{{\bf Iteration training parameter:}}
As shown in Algorithm~\ref{alg:interCNNtraining}, the proposed training strategy is to train interCNN for a fixed $K$ number of iterations per batch. Meaning the predictions of every batch of images are iteratively updated together with their respective scribbles and fed back into interCNN for $K$ number of consecutive iterations before moving on to the next batch. To inspect the influence of number of iterations during training, we varied $K$ from 1 to 15. 

\begin{figure}[tbh]
\centering
\includegraphics[trim = 0mm 0mm 0mm 0mm, clip,height=0.223\textheight]{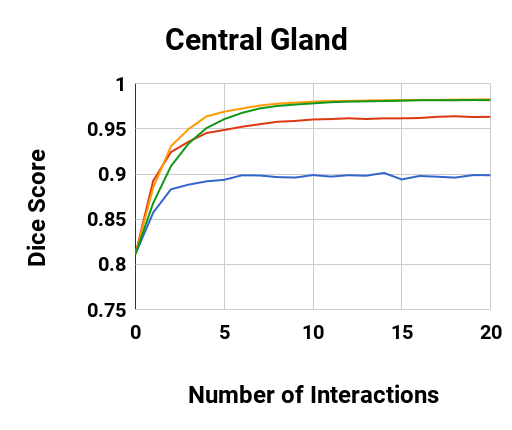}
\includegraphics[trim = 0mm 0mm 0mm 0mm, clip,height=0.223
\textheight]{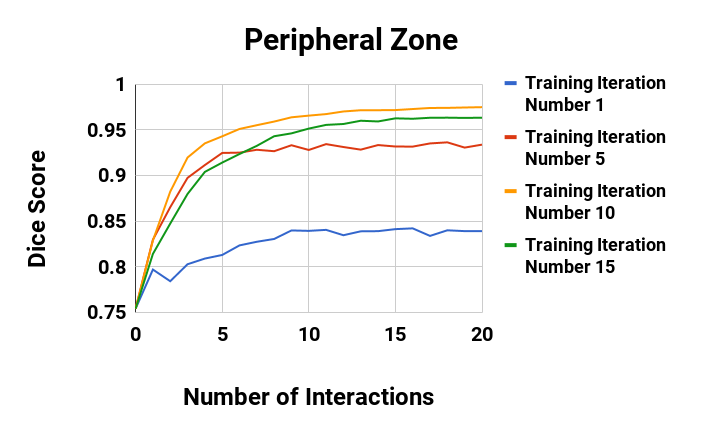}
\caption{Segmentation performance for interCNN trained with 1 to 15 iterations ($K$) for (left) central gland and (right) peripheral zone.}
\label{fig:influenceOfIterations}
\end{figure}

The results for the two prostate structures are shown in Fig.~\ref{fig:influenceOfIterations}.
It can be seen that the Dice score improvement is substantially lower if iteration parameter $K$ is set to 1 or 5, compared to a $K$ of 10 and higher. Even though there is an initial improvement of the Dice score with a low $K$, the improvement slows down at later interaction iterations. One possible explanation for this observation could be that the interCNN is mostly confronted with large incorrectly classified areas during training for low iteration parameters and learns to make large segmentation adjustments which is not required or beneficial at later stages. 
%A second interesting observation is that the difference of the Dice score improvement is small for iteration parameters of 10 and larger. In addition, it can be seen that the Dice score is not improved substantially after 10 iterations for high iteration parameters. This implies that there won’t be much learning potential for the interCNN with an iterative parameter larger than 10. 
%\todo{[CT: the second interesting observation is included in the first observation. The number of incorrect pixels is decreasing, leading to a saturation effect of the learning as performance reaches its limit. Does it attain the upper limit given by training it on full segmentations?]}
%

%\subsection{Comparison to interactive segmentation from scratch}
\paragraph{{\bf Comparison to interactive segmentation from scratch:}}
To evaluate the value of segmentation editing compared to state-of-the-art interactive segmentation from scratch, we looked at two recently proposed approaches. 
%To investigate how our algorithm compared to state-of-the art methods we looked at the following two approaches from literature.

%{\bf GrabCut}: This is a well established traditional interactive segmentation algorithm based on a Gaussian Mixture Model (GMM)~\cite{rother2004grabcut}. It is initialized by a bounding box the user has to draw around the object of interest and the segmentation is further updated based on the input scribbles from an user. We used the openCV [ref] implementation of this algorithm to obtain our test results.

{\bf UI-Net}: The method is based on a CNN taking scribbles and the image as input to update its segmentation~\cite{amrehn2017ui}. No automatic segmentation takes place, but rather initial scribbles are provided by the user. In contrast to~\cite{amrehn2017ui}, the initial scribbles were chosen randomly and not by erosion and dilation.
As CNN we used the same U-Net architecture as for interCNN.

{\bf BIFSeg}: This method is based on fine-tuning the last-layer of a CNN to update segmentations based on user inputs~\cite{wang2018interactive}. The algorithm starts by asking the user to draw a bounding-box around the object of interest. An initial segmentation is then computed and the scribbles of the user in the following iterations are used to fine-tune the last layer of the CNN that predicted the initial segmentation. We used their open-source code to benchmark against, which is claimed to also work on objects not seen during training.

\begin{figure}
\floatbox[{\capbeside\thisfloatsetup{capbesideposition={right,center},capbesidewidth=0.33\textwidth}}]{figure}[\FBwidth]
{\caption{Segmentation performance of proposed method (interCNN) in comparison to state-of-the-art methods for increasing number of user interactions (1-20).}\label{fig:comparisonWithLiterature}}
{\includegraphics[trim = 0mm 0mm 0mm 20mm, clip,width=0.63\textwidth]{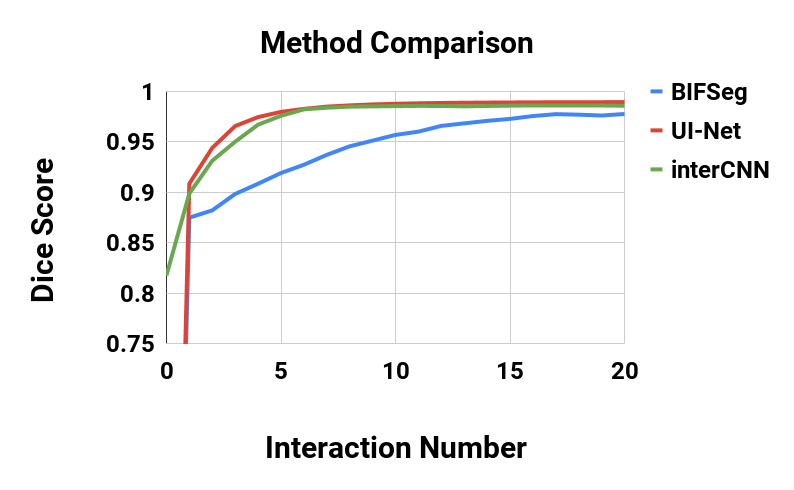}}
\end{figure}

In Fig.~\ref{fig:comparisonWithLiterature} the results of the comparison to interCNN with 10 training iterations can be seen. As both of the methods we compare to require user interaction, their Dice scores only start at iteration one. For BIFSeg this initial input is the bounding-box annotation. We investigated how the Dice score changed for all these methods over the course of 20 user interactions.
It can be observed that interCNN, which edits existing segmentations, required substantially fewer user interactions than BIFSeg to reach a high Dice score (5 vs. 20). The performance of UI-Net, on the other hand, was very similar to the proposed method for this dataset, but it also used the full training dataset as it was trained from scratch.

\begin{figure}[tbh]
\centering
\includegraphics[trim = 0mm 25mm 0mm 24mm, clip,height=0.1\textheight]{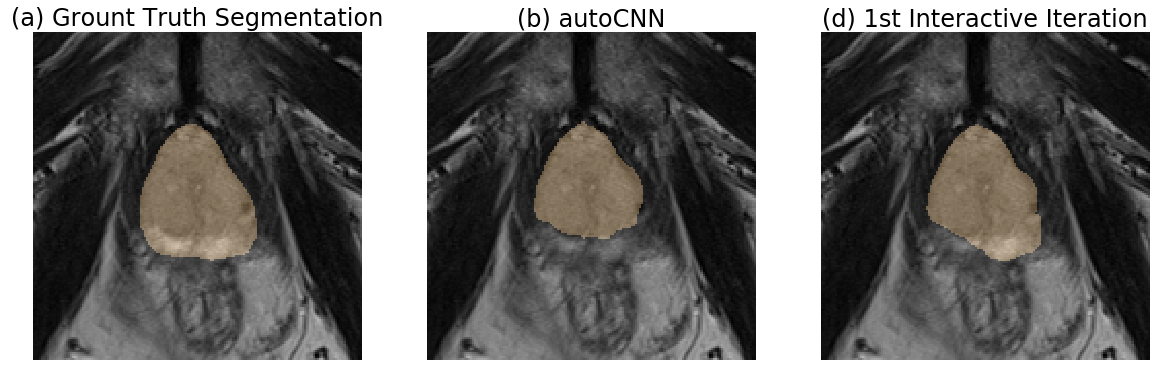}
\includegraphics[trim = 280mm 25mm 0mm 24mm, clip,height=0.1\textheight]{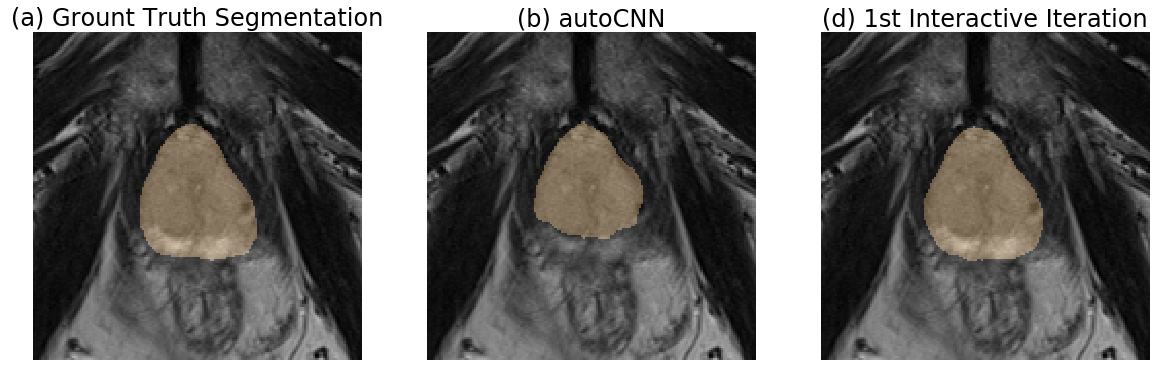}
\caption{Visual examples: segmentation overlays for (left$\rightarrow$right) ground truth, autoCNN (DSC:0.84), and interCNN after interaction 1 (DSC:0.93) and 5 (DSC:0.98).}
\label{fig:exampleResults}
\end{figure}

%
% \begin{figure}[tbh]
% \centering
% \includegraphics[trim = 0mm 0mm 0mm 0mm, clip,width=0.47\textwidth]{Method_Comparison.png}
% \caption{Segmentation performance of proposed method (interCNN) in comparison to state-of-the-art methods for increasing number of user interactions (1-20).}
% \label{fig:comparisonWithLiterature}
% \end{figure}
%
The iterative improvement of the base segmentation by interCNN is illustrated on a representative test example in Fig.~\ref{fig:exampleResults}.

\section{Conclusions}

We proposed an iterative interaction training strategy for efficient segmentation editing with networks. %Efficient training of the interactive method was possible due to a scribble generating robot-user and \addct{task-specific} iterative training. 
Compared to non-iterative training, the proposed strategy yielded higher segmentation accuracy. 
The difference was the highest when the iteration parameter for training was at ten and higher. 
%We saw that the optimal iteration parameter for \addct{training} the method was at ten and higher. 
The proposed strategy allows the CNN to learn to correct small and large errors.
Finally, we compared our method to alternatives that perform interactive segmentation from scratch. 
We observed that interCNN when trained with the proposed strategy yielded results on par with the state-of-the-art methods. 
%and observed that \addct{fewer user edits were required with our proposed method} \chct{superior correction convergence time was achieved with our implementation}. 
The advantage of segmentation editing networks, such as interCNN, compared to interaction segmentation from scratch is that they do not need user interaction to initialize segmentation. 

%\section{Acknowledgment}
%\todo{Is there a standard text for this section? CT: for anonymous papers no acknowledgments needed to avoid revealing something}
\noindent {\bf Acknowledgments:} We thank the Swiss Data Science Center (project C17-04 deepMICROIA) for funding and acknowledge NVIDIA for GPU support.

\bibliographystyle{splncs03}
\bibliography{refs}

\end{document}